\documentclass[runningheads]{llncs}
\usepackage{graphicx}
\usepackage{algorithm}
\usepackage{algorithmic}
\usepackage{amsmath}
\usepackage{textcomp}
\usepackage{subfig}
\usepackage{multirow}
\begin{document}
\title{Reinforcement Learning with Expert Trajectory For Quantitative Trading}


\author{Sihang Chen \and Weiqi Luo \and Chao Yu}

\institute{School of Computer Science and Engineering\\Sun Yat-Sen University\\Guangzhou, China\\
\email{chensh236@mail2.sysu.edu.cn,\\ \{luoweiqi, yuchao3\}@mail.sysu.edu.cn}}

\maketitle              
\begin{abstract}
In recent years, quantitative investment methods combined with artificial intelligence have attracted more and more attention from investors and researchers. Existing related methods based on the supervised learning are not very suitable for learning problems with long-term goals and delayed rewards in real futures trading. In this paper, therefore, we model the price prediction problem as a Markov decision process (MDP), and optimize it by reinforcement learning  with expert trajectory. In the proposed method, we employ more than 100 short-term alpha factors instead of price, volume and several technical factors in used existing methods to describe the states of MDP.  Furthermore, unlike DQN (deep Q-learning) and BC (behavior cloning) in related methods, we introduce expert experience in training stage, and consider both the expert-environment interaction and the agent-environment interaction to design the temporal difference error so that the agents are more adaptable for inevitable noise in financial data.  Experimental results evaluated on share price index futures  in China,  including IF (CSI 300) and IC (CSI 500), show that the advantages of the proposed method compared with three typical technical analysis and two deep leaning based methods. 
\end{abstract}

\section{Introduction} 
Quantitative investment analysis \cite{guida2019big} is the use of mathematical and statistical methods to assist investors in making profitable investment decisions.  With the help of high-performance computer technology \cite{High-Frequency},  we  can  effectively  analyze  huge amounts of financial data  in a short  time,  and automatically execute orders according to the pre-programmed instructions.   Recently,   quantitative  investment methods  have attracted more and more attention from investors and researchers.

Fundamental analysis and technical analysis are the  two primary methods used to analyze  financial data and make investment decisions \cite{schwager1984complete}. Fundamental analysis  determines the fair value of the business via analyzing the company's financial statements.  While the  technical analysis  \cite{murphy1999technical} is a methodology for forecasting the direction of prices via analyzing  past market data, primarily price and volume.   For instance,  Moving Average Convergence / Divergence (MACD) \cite{appel2008understanding} employs moving averages to reveal changes in the strength, direction, momentum, and duration of a trend in a contract price;  Dual Thrust strategy \cite{pruitt2012building} firstly generates two lines, i.e., BuyLine and SellLine, according to the mean and variance values during a period of time, and then generates the trading signal when the current price breaks through the BuyLine or SellLine.  The above mentioned models are relatively simple,  and thus they are widely used as baselines in quantitative analysis.  However,  their generalization ability \cite{DingWSGG20} is relatively weak.  

The generalization ability of quantitative method  can be improved to a certain extent by combing with some advanced supervised learning techniques.  For instance, the method  \cite{chen2015lstm} feeds the historic price data, technical analysis data and economic fundamentals into LSTM (Long Short-Term Memory) \cite{schmidhuber1997long}  to predict the price fluctuation in daily frequency;   the method  \cite{FengC0DSC19} uses an attentive LSTM model to predict the movement of contract price in the near future and employs adversarial training to improve the generalization of the model;  the method  \cite{0101BHCXS20} uses a LSTM model with graph convolutional network to process overnight news, and then predicts the overnight stock movement between the previous close price and the open price.  However,  the accuracy of the forecast can only reflect part of the model performance \cite{li2019deep},  long-term goals and delayed rewards should be considered in real futures trading.  In addition,  several important factors (e.g., transaction cost) in real transaction  are not considered at all in these methods.   

Recently,  some quantitative methods based on Reinforcement Learning (RL)  \cite{sutton2018reinforcement} have been proposed to solve sequential trading decision-making problems,  and aim to  maximize the expectation of cumulative reward.   For instance, 
the method \cite{deng2016deep} firstly employs fuzzy network to reduce the uncertainty of the input data and employs auto-encoder  to reduce the dimension of the features,  and then uses a recurrent neural network to map the state in the environment to agent's action directly;  the methods \cite{tan2011stock,zhang2020deep} use the deep Q-network (DQN) \cite{tan2011stock} to approximate the action-value function in order to estimate the value of agent's trading decision;  the method \cite{si2017multi} sets multiple targets according to the mean and variance of return in financial market to balance the profits and risks.   
Note that the above reinforcement learning  based methods do not depend on any prior knowledge,  thus they often require many experiences for learning,  preventing them from being practical in most real situations \cite{BrysHSCTN15}. 
Imitation learning is a learning pattern in RL, which introduces prior knowledge in the training stage.
In this pattern, an expert demonstrates how to solve the task, and the agent imitates these demonstrations to select the corresponding actions. Behavior cloning (BC) \cite{BC1,BC2} is a typical method in imitation learning. 
Until now, there are just a few related works in the financial field.  For instance, the method \cite{liu2020adaptive} firstly constructs an actor-critic network to map the state to the action, and then uses behavior cloning method to pre-train this network. 
However, BC based methods use supervised learning to imitate expert's demonstrations greedily instead of reasoning about the consequences of actions.  Thus, when the agent drifts and encounters out-of-distribution states, the agent does not know how to return to the demonstrated states \cite{reddy2019sqil}. 

In this paper,  we firstly model the index prediction problem as a Markov decision process, and then  propose a RL based method with expert trajectory for quantitative trading.  Unlike behavior cloning that sets the expert actions as demonstrations and uses supervised learning to imitate expert's actions,  we design the  temporal-difference error (TD-error) derived from  both the agent-environment interaction and the expert-environment interaction.  In this way,  the agents are more adaptable for inevitable noise in financial data.  In addition, in order to  express the current situation of the market more comprehensively,  we introduce more than 100 short period alpha factors as the stage instead of several technical indicators  used in related work.  Experimental results evaluated on share price index futures  in China,  including IF (CSI 300) and IC (CSI 500),  show the advantages of the proposed method compared with three typical technical analysis, including Buy \& Hold, MACD and Dual Thrust,  and  two RL based methods, including DQN and BC. 
 
The rest of this paper is arranged as follows.
Section \ref{sec:Preliminaries} shows the preliminaries of RL used in the proposed method.
Section \ref{sec:Proposed_work} shows the proposed method;
Section \ref{sec:Experiments} shows the comparative results and discussions.
Finally, the concluding remarks of this paper and future works are given in Section \ref{sec:Conclusion}. 

\section {Preliminaries}
\label{sec:Preliminaries}

In this section, we will describe some preliminaries about RL  \cite{sutton2018reinforcement}  used in the proposed method, such as MDP,  Q-learning,  $\epsilon-$greedy and experience replay.  

\paragraph{MDP:}In a finite  MDP,  the agent and environment interact at each discrete time step.  At the time step  $t$,   the agent first receives some representation  $s_t  \in \mathcal{S}$ of the environment's state,  and then selects an an action  $a_t \in \mathcal{A}$ according to  $s_t$.   Then,  the agent gets a numerical reward  $r_t  \in \mathcal{R}$ from the environment, and then the agent turns to a new stage $s_{t+1}$.     
The interactions between the agent and the environment would finally produces a trajectory 
$\tau = [s_0, a_0, r_0, s_1, a_1, r_1,  \ldots ]$.

\paragraph{Q-learning:}  RL aims to find an optimal policy $\pi:\mathcal{S}\rightarrow\mathcal{A}$ to maximize  the cumulative reward $G_t$ of agent: 
\begin{equation}
G_t =\sum_{k = t}^{T} \gamma ^k r_{t + k}
\end{equation}
where $\gamma \in [0, 1]$ is the discount factor;  $T$ is a final time step in the current episode. 
 
The value function describes the expectation of the cumulative reward, and the optimal policies share the same optimal value function. 
Q-table expresses the maximum of value function which is calculated from each state and each action.
Q-learning \cite{watkins1992q} updates the Q-table based on the following formula until convergence,
\begin{equation}
    Q(s_t, a_t) \leftarrow Q(s_t, a_t) + \alpha [r_t + \gamma max_a Q(s_{t+1}, a) - Q(s_t, a_t)]
\end{equation}
where $ \alpha $ is the learning rate.  
Q-network is used to approximate the Q-table which is too large, and it outputs $Q(\cdot, \cdot; \theta)$, where $\theta$ denotes the parameters in Q-network.

\paragraph{$\epsilon$-greedy:}   $\epsilon-$greedy \cite{DBLP:TokicP11} aims to increase agent's exploration ability. 
At each time step $t$ in the training stage, the agent chooses the action randomly with probability of $\epsilon$.
With the probability of $1-\epsilon$, the agent selects the action with the largest action value, denoted as,
\begin{equation}
a_{t} = \left\{
    \begin{aligned}
    &\max_{a}Q(s_t, a;\theta), \ \text{with\ probability\ }(1 - \epsilon) \\
    &\text{a\ random\ action}, \ \text{with\ probability\ }\epsilon
    \end{aligned}
    \right.
\end{equation}
Note that  $\epsilon$ decreases gradually during the training stage.

\paragraph{Experience replay:} Experience replay \cite{DBLP:Lin92} aims to break the temporal correlation among MDP samples and reduce the amount of experience required to learn. After the interaction between the agent and the environment, the current sample is saved into the replay buffer. When the buffer is full, we randomly select a mini-batch of samples from the buffer to calculate the loss function.

\section {Proposed Method}
\label{sec:Proposed_work}
In this section, we first describe the problem definition of MDP in futures trading, and then we show the details of the proposed method based on RL with  expert trajectory.  

\subsection{Problem Definition}
\label{sec:Problem_definition}
In the proposed method, we model the trading-making problem as a finite MDP. Thus, the discrete probability distribution of  the state $s_t$  and the reward $r_t$ just depends on the last state and action.  In the following, we will describe what are the state set  $\mathcal{S}$, action set $\mathcal{A}$ and the reward design.

 \begin{itemize}
\item \textbf{State Set  $\mathcal{S}$:}  At each time step $t$, the  representations $s_t \in \mathcal{S}$ of the  environment's state include  historical market indicators, such as OHLC (open, high, low, close) prices, volume and amount, and over 100  factors \footnote{https://www.joinquant.com/data/dict/alpha191} calculated from those fundamental data such as volatility, momentum and other statistics over a period of time.  
 
\item \textbf{Action Set $\mathcal{A}$:} For a fair comparison with other trading methods,  we assume that the  transaction at each time step is one unit.  Therefore,   the action set  is defined as $\mathcal{A} = \{-1, 0, 1\}$, where $-1$ denotes  short position;   $0$ denotes no holding;   $1$ denotes long position.

\item \textbf{Reward Design in Training and Testing stages: } Since the  expert trajectory is introduced in the proposed method,  there is no need for the agent to learn from reward signal.  Like  \cite{reddy2019sqil},  in the training stage,  the reward of the expert is fixed as $r = 1$  and the reward of the agent is fixed as $r = 0$. 
In  the testing stage, however, the real reward of the agent is used to measure the performance of the proposed method \cite{deng2016deep}.   At the time step $t$,  the agent gains the reward: 
\begin{equation}
r_t = a_{t-1}(p_t - p_{t-1}) - c|a_t - a_{t-1}|
\end{equation}
where $a_t$ is the current action,  $a_{t-1}$ is the previous action,  while  $c$ is rate of transaction cost.  Note that when the action is unchanged, i.e.,  $|a_t - a_{t-1}| =0$,  the agent maintains current position without any transaction cost.  When the action changes, i.e.,   $|a_t - a_{t-1}| = 1$ or 2, the corresponding  contraction cost would be $c$ and $2c$. 
\end{itemize}
\begin{figure*}
    \centering
    \includegraphics[scale=0.45]{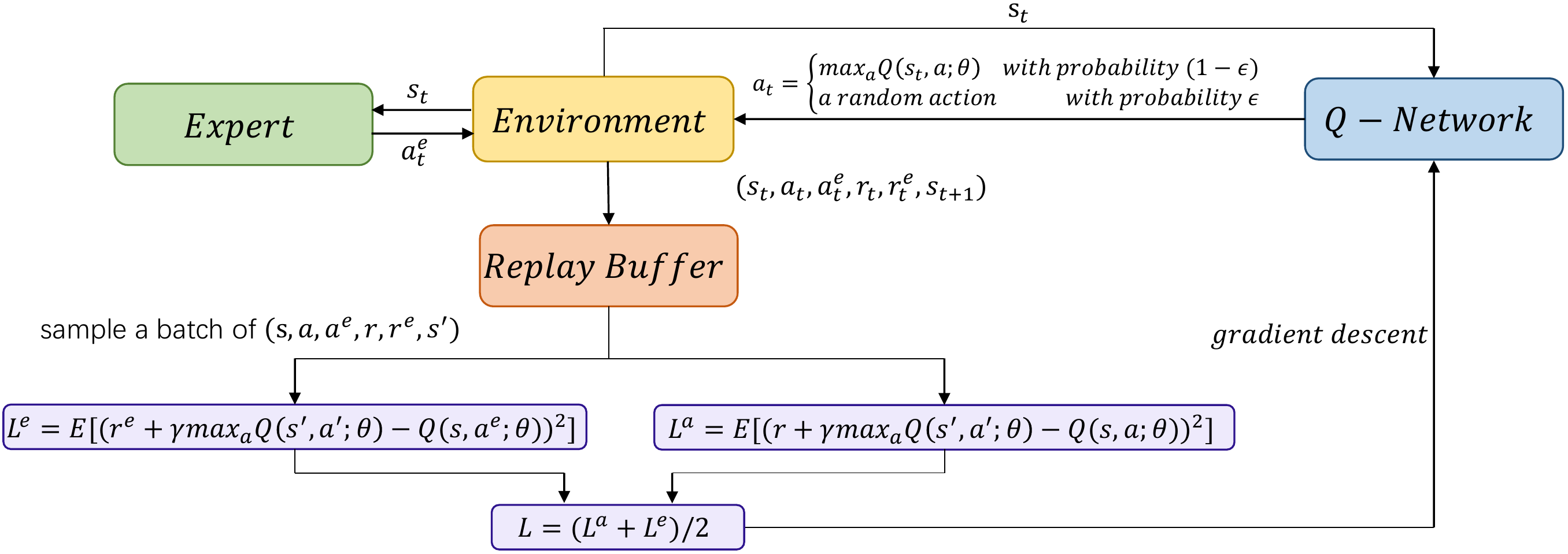}
    \caption{The framework of the proposed method}
    \label{fig:framework}
 \end{figure*}
\subsection{The Proposed RL with expert trajectory}
\label{subsec:proposed}
As illustrated in Fig. \ref{fig:framework},  there are four main components in the  proposed framework,  that is,  Expert,  Q-Network,  Environment and Reply Buffer. In the following, we first describe the four components respectively, and then describe the training stage in the proposed method.  

\begin{itemize}
\item \textbf{Expert:}  In the training stage, the expert can always take correct trading actions and generate the corresponding expert trajectory as the demonstration.  According to the demonstration, the agent optimizes the policy via imitating the expert's actions.  Usually,  the expert actions can  shorten the inefficient random exploration stage. 

As described in section \ref{sec:Problem_definition}, we assume that at each time step $t$,  investor should take an actor from the actor set, i.e.,  short, no holding, and long,  and the transaction is one unit.
Thus,  at each time step $t$,  the investment profit  only depends on the close price at the next time step $t+1$.
If the investor can make the correct action at every time step,  the accumulated profit would be the highest. 
In other words,  we can obtain the optimal policy using Greedy algorithm if we can  always predict the close price at the next time step correctly.  In the investigated  problem,  therefore, the expert trajectory is  simple,  which is composed of the correct trading actions during the training:  taking a long position when the close price increases from the time step $t$ to $t+1$;   taking a short position when the close price decreases;  no holding when the close price does not change.  
 
\item \textbf{Q-network:} Q-Network is composed of a LSTM model and a full-connected (FC) layer,  and it aims to approximate the state-action value function.  The input of the Q-network is a given state that consists of a sequence of historical market indicators and factors, and the output is the action value tensor for each possible actions.

\item \textbf{Environment:} Environment is a stimulated financial market.  The agent acts as an investor to interact with the environment.  Agent takes a trading action according to a given stage in the  environment,   and then obtains the reward from the environment and observes the next state.   
    
\item \textbf{Replay Buffer:}  Replay Buffer is used  to achieve experience replay, which aims to  reduce the temporal correlation among MDP samples.  At each time step, the experiences of agent and expert are saved to replay buffer. When the buffer is full, a mini-batch of samples are randomly selected to update the weights in Q-network.   
\end{itemize}

At each time step in the training process,  the agent firstly observes the state $s_t$,  and then takes an action $a_t$ according to the output of the Q-network  and the $\epsilon$-greedy rule.  And then  the agent obtains reward $r_t$ from the environment and observes the next state $s_{t+1}$.   At the same time,  the agent obtains the demonstrative action $a_t^e$ and the expert reward $r_t^e$ from expert trajectory.  To achieve experience replay, the sample $(s_t, a_t, a_t^e, r_t, r_t^e, s_{t+1})$ is saved into the replay buffer.
The above process  repeats until the replay buffer is full.

When the the replay buffer is full,  we randomly select a mini-batch of samples $(s, a, a^e, r, r^e, s')$  from the buffer,  where $s$ denotes the current state, and $s'$ denotes the next state.   $a$ ($a^e$) and $r$ ($r^e$) denote agent's (expert's) action and reward respectively. 
Then the agent loss $L^{a}$ and expert loss $L^{e}$ are defined as the  TD-error generated from the agent-environment and expert-environment respectively, that is 
\begin{equation}
L^{a} = E[(r + \gamma\cdot max_{a}Q(s', a; \theta) - Q(s, a; \theta))^2]
\end{equation}
\begin{equation}
L^{e} = E[(r^e + \gamma\cdot max_{a}Q(s', a; \theta) - Q(s, a^e; \theta))^2]
\end{equation}
 
\vspace{0.5em}Finally, the loss function  $L$ is defined as the average value of  $L^{a}$ and   $L^{e}$: 
\begin{equation}
L = \frac{L^{a} + L^{e}}{2}
\end{equation}

We use $L$ to calculate the gradient of weights in Q-network via back propagation.
The pseudo-code of the training process is shown in Pseudo-code \ref{alg:algorithm}.

 \begin{algorithm}[tb]
    \caption{Reinforcement learning with expert trajectory}
    \label{alg:algorithm}
    \textbf{Input}:\\
    Episode number $N$; Time step number $T$ in an episode;\\
    Buffer capacity $N_c$; Batch size $N_s$;\\
    Actions in expert trajectory  $[a_{0}^{e}, a_{1}^{e}, a_{2}^{e},\dots]$.\\
    \textbf{Parameter}: \\ Q-Network parameters $\theta$; 
    \begin{algorithmic}[1] 
    \STATE $episode\leftarrow 0$;
    \WHILE{$episode < N$}
        \STATE $t \leftarrow 0$;
        \STATE Observe initial state $s_0$;
        \WHILE {$t < T$}
            \STATE $b_c \leftarrow 0$;
            \WHILE{$b_c < N_c$ \& $t < T$ }
                \STATE Select a random action with probability $\epsilon$ or select an action with max action-value:\\ $a_t \leftarrow max_{a}Q(s_t, a; \theta)$ from Q-network;
                \STATE Interact with the environment, obtain the reward $r_t$ and observe the next state $s_{t+1}$;
                \STATE Take the expert action $a_t^{e}$ to interact with the environment, obtain the reward $r^{e}_t$ and observe the next state $s_{t+1}$; 
                \STATE Save the resulting sample $(s_t, a_t, a_t^e, r_t, r_t^e, s_{t+1})$ into replay buffer;
                \STATE $t \leftarrow t+1$;
                \STATE $b_c \leftarrow b_c+1$;
            \ENDWHILE
            \STATE Select a mini-batch of sample $(s, a, a^e, r, r^e, s')$ with the size $N_s$ randomly from the buffer;
            \STATE Calculate loss function: \\$L = (L^{a} + L^{e}) / 2$;
            \STATE Use $L$ to update $\theta$ in Q-network;
            \STATE Clear replay buffer; 
        \ENDWHILE
        \STATE $episode \leftarrow episode+1$;
    \ENDWHILE
    \end{algorithmic}
    \end{algorithm}

\section{Experimental Results and Discussions}
\label{sec:Experiments}
In this section, we first describe the experimental setup used in our experiments, and then give a brief  description on metrics and the baseline methods.  Finally, we show the comparative results  and discussions.  
 
\subsection{Experimental Setup}
\label{subsec:set}
Two major futures indexes in China,  e.g., IF (CSI 300 index) and IC (CSI 500 index), are included in our experiments.
We obtain the tick level data including the price, volume and transaction amount from Oct. 1st, 2015 to Oct. 15th, 2020.  The data from Oct. 1st, 2015 to Oct. 1st, 2019 is used in the training stage, while the remaining data is used for testing.   The original tick level data is aggregated into 5-min lever data for calculating the factors in our experiments. 

Note that all experimental data is from Tinysoft \footnote{\url{http://www.tinysoft.com.cn/}},  a professional financial data platform in China. In this  platform, the contract with the largest volume in the previous day is considered as the dominant contract of this commodity on the next day.  Thus,  day trading strategy is used for all test methods, meaning that all positions are closed before the market closes in order to avoid unmanageable risks and negative price gaps between two different trading days.
 In addition,  we set the  discount factor  $\gamma=0.992$, and set the transaction cost $C=0.023$ \textperthousand  \hspace{0.2em}according to the regulation of financial futures exchange, and  the margin system is not considered.

In the testing stage,  a stop-loss  strategy is used in the proposed method.  For each time step $t$,  the strategy first calculates the mean $\mu_{k}$ and standard deviation $\delta_{k}$ according to the price fluctuation at previous $k$ time steps.  
If the price breaks through upper bound, i.e.,  $\mu_{k}+\delta_{k}$  while we hold a short position,  or the price breaks through the lower bound, i.e., $\mu_{k}-\delta_{k}$  while we hold a long position,  then the position is closed.  According to our experiments (Refer to section \ref{sec:comparative} for details),  we fixed $k=25$.

\subsection{Evaluation Criteria} 
 To evaluate the profit and risk aversion ability of the proposed model, three commonly used criterias are employed.
 
\begin{itemize}
\item \textbf{Accumulated profits:} The accumulated profits of the model is expressed as the sum of each time step during the testing stage, i.e. $R = \sum_{t = 0}^{T} r_t$, while $T$ is the number of time steps in the test set.
   
\item \textbf{Sharp ratio:} Sharp ratio is used to measure profits expectations and the proportion of risk.
    Sharpe ratio can be expressed as $Sharp = \frac{E(r)}{\sigma(r)}$, where $E(r)$ is the expectation of all the benefits of the test set and the standard deviation of the profits is $\sigma(r)$.
    
 \item \textbf{Sortino ratio:}  Sortino ratio is similar to sharp ratio, but it uses the downward standard $\sigma(r_d) $, where $r_d$ represents downward earnings (compound earnings). The sortino ratio can be expressed as $ Sortino = \frac{E(r)}{\sigma(r_d)} $.
\end{itemize}

\subsection{Baseline Methods}
\label{subsec:baselines}
Three typical technique analysis, i.e. Buy \&  Hold, MACD \cite{appel2008understanding}, and Dual Thrust \cite{pruitt2012building}, and two RL based methods, i.e., DQN and BC,  are included for comparative studies in our experiments. 
\begin{figure*}[!t]
\centering
\subfloat[Compared with three technique analysis] {\includegraphics[width =8.7cm]{{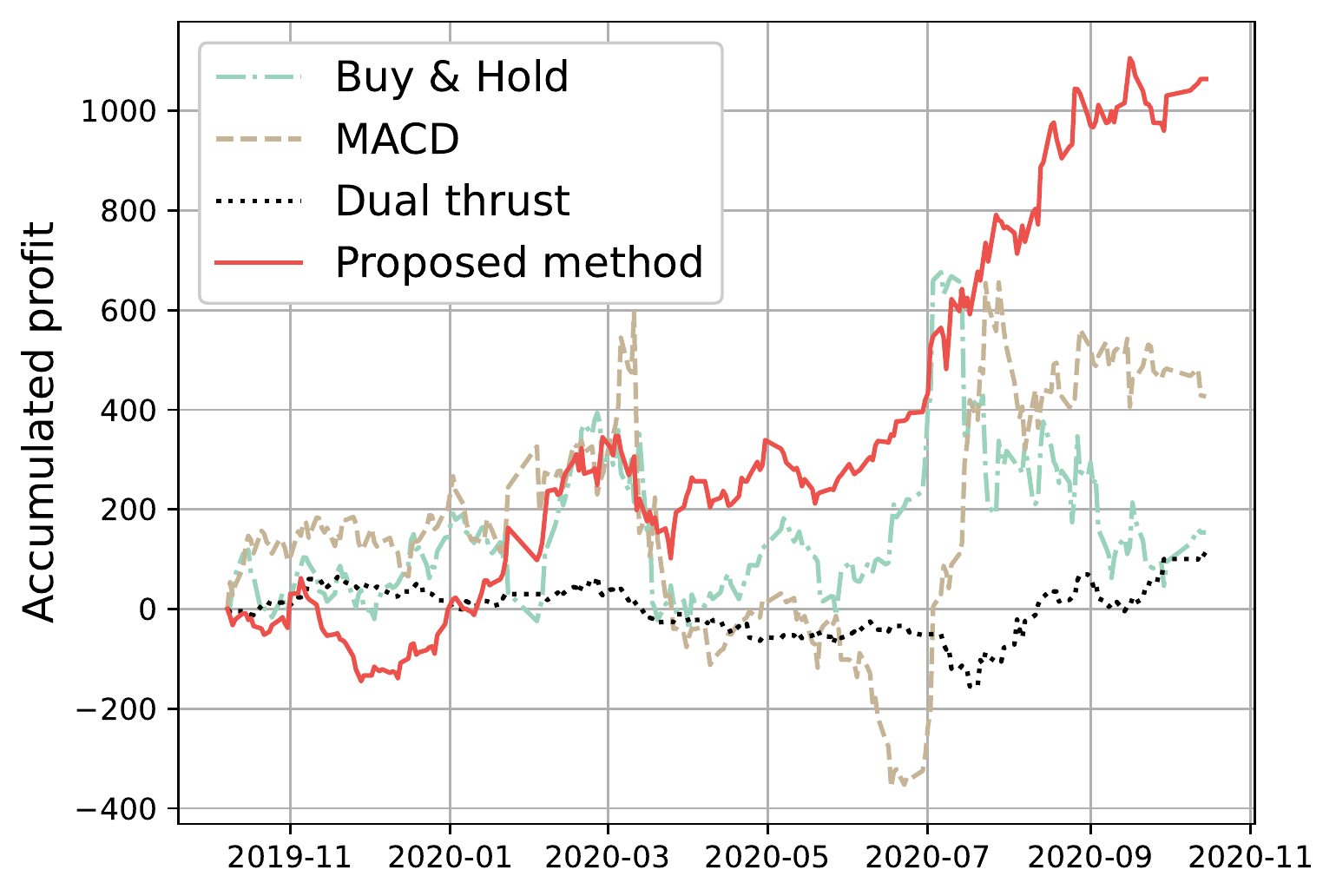}}}  \quad
\subfloat[Compared with two RL related  methods] {\includegraphics[width=8.7cm]{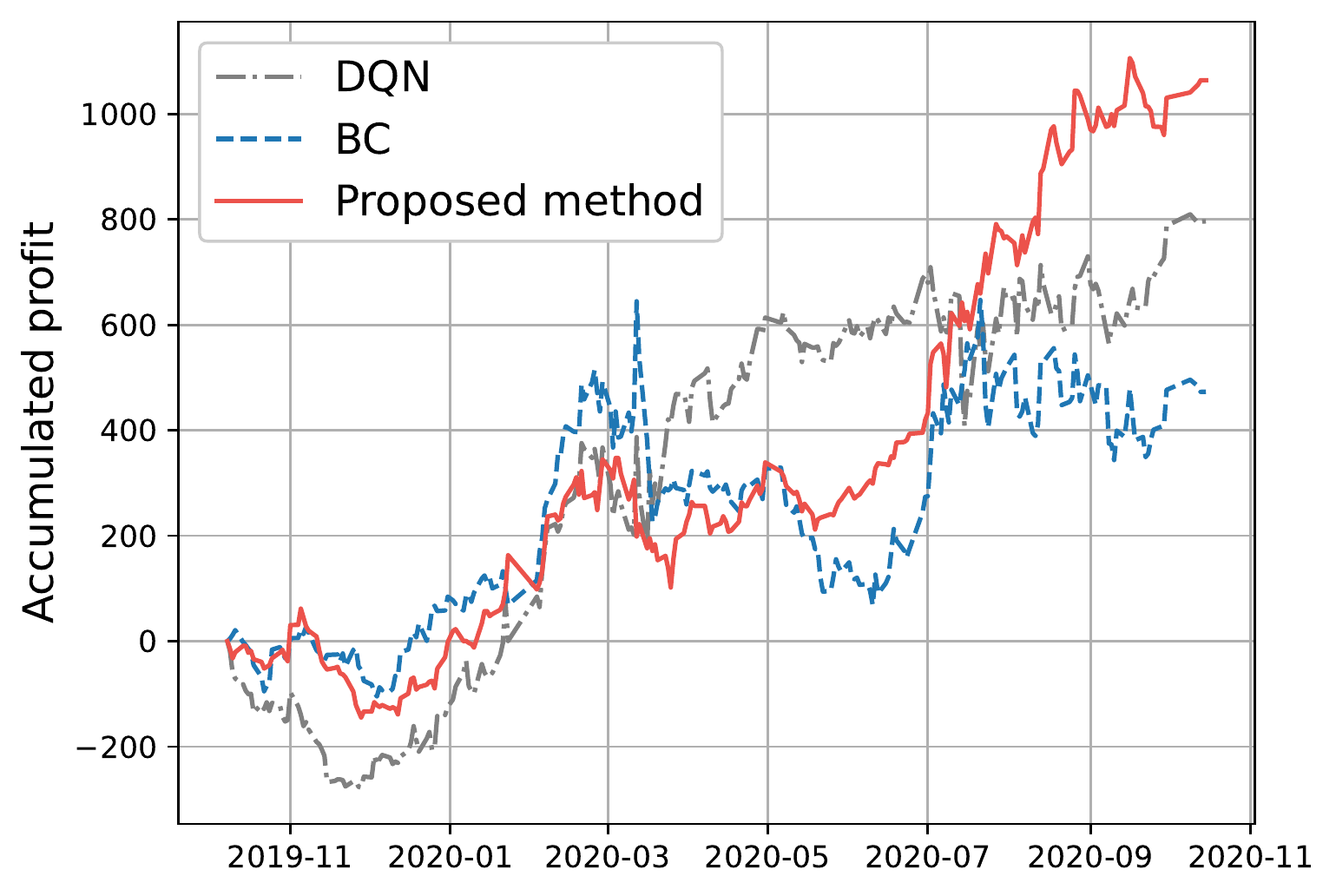}}  
\caption{The  profit curves of various methods during the testing date from Oct. 2, 2019 to 
Oct. 15, 2020. (a) Compared with three technique analysis, including Buy \& Hold,  MACD, and Dual trust; (b) Compared with two RL based methods, including DQN and BC.}
\label{IF}
\end{figure*}
\begin{itemize}
\item \textbf{Buy \& Hold: }  Investor buys a contract at the beginning and  holds  it in the whole day of trading.  Then the position is closed at the end of the trading day. 

\item \textbf{MACD:} This method includes the DIF line and the DEA line. The DIF line is the difference between a short period exponential moving average (EMA) and a longer period EMA of the pr ice series.
The DEA line is the weighted moving average of the DIF line.
When the  $DIF$ line breaks through $DEA$ line upwards and $DIF$ value is bigger than 0, long signal is generated, and if $DIF$ line breaks through $DEA$ line downwards and $DIF$ value is smaller than 0, short signal is generated.

\item \textbf{Dual Thrust:}  Let $R = max (HH - LC, HC - LL) $, where 
$LC$  is the minimum of close price, $HC$ is the maximum of close price, $LL$ is the minimum of the lowest price, and $HH$ is the maximum of the highest price  during a period of time. Using the sum and difference between $R$ and open price, we obtain the BuyLine and SellLine respectively. When the price breaks through BuyLine/SellLine, the long/short signal is generated in Dual Thrust.

\item \textbf{DQN:} Compared with the proposed method,  the DQN based method trains the agent through the reward, which is designed as the actual profit,  instead of the expert trajectory.  The loss is defined as the TD-error between two adjacent time steps.
Then, the other modules remain unchanged, and the stop-loss strategy is attached.

\item \textbf{BC:} Behavior cloning is another way of imitation learning.
In BC, the actions in expert trajectory are used as labels and also the Q-Network is remained.
While training, the cross entropy is used to measure the gap between the actions generated by the Q-Network and labels, with an Adam optimizer to reduce the loss, and the stop-loss strategy is attached.
\end{itemize} 

\begin{table}
\caption{The overall performances of various approaches in IF and IC markets.  Those values with an asterisk (*) denote the best results in the corresponding cases. }
\label{tab:performance}
\begin{center}
\begin{tabular}{c | c | c c c}
            \hline
            \multirow{2}{*}{\textbf{Market}} & \multirow{2}{*}{\textbf{Approach}} & \multicolumn{3}{c}{\textbf{Performance}}\\
            \textbf{}&\textbf{}&\textbf{Profits}&\textbf{Sharp}&\textbf{Sortino}\\
            \hline
            \multirow{2}{*}{{IF}} & {Buy \& hold}&153.8 & 0.2 &0.28\\
            \textbf{} & {MACD}&426.6&0.59&0.55\\
            \textbf{} & {Dual thrust}&107.2&0.57&0.29\\ 
            \textbf{} & {DQN}&796.4&1.25&1.84\\
            \textbf{} & {BC}&473.2&0.74&1.08\\
            \textbf{} & {Proposed}&\textbf{1064.0 *} &\textbf{2.09 *} & \textbf{2.47 *}\\
            \hline
            \multirow{2}{*}{{IC}} & {Buy \& hold}&545.5&0.45&0.62\\
            \textbf{} & {MACD}&55.6&0.05&0.04\\
            \textbf{} & {Dual thrust}&179.0&0.4&0.16\\
            \textbf{} & {DQN}&1317.3&1.27&1.85\\
            \textbf{} & {BC}&-113.8&-0.11&-0.15\\
            \textbf{} & {Proposed}& \textbf{1934.3 *} &\textbf{1.86 *}& \textbf{2.86 *}\\
            \hline
		\end{tabular}
		\label{comparation}
	\end{center}
\end{table}

\subsection{Comparative Studies}
\label{sec:comparative}

In this section, we will compare the proposed method with other five related works as described previously in section \ref{subsec:baselines}. For the IF market,  the accumulated profits of various methods during the testing date (i.e.,  from Oct. 2, 2019 to Oct. 15 2020)  are shown in Fig. \ref{IF}.  From Fig.  \ref{IF}, we observe that the curve of the proposed method is higher than those of others methods in most cases,  and is always the highest at the end of testing period, which means that the proposed method  gains more profits stably.  

The overall performances  among the testing date for both IF and IC markets are  shown in Table \ref{tab:performance}.  From Table \ref{tab:performance}, we observe the two following observations:

\begin{itemize}
\item First of all, the proposed method always achieves the best performance on the three  criteria both for IF and IC markets.  Taking IF for instance, we achieve as high as 1064, and both the sharp and sortino rates are higher than 2.0.  Except for DQN,   the profits of other methods are less than 474,  and their  sharp and sortino rates are less than  1.10. 

\item For the three typical technique analysis, they can always achieve the positive profits,  although  their profits are not very high. For the DQN, it achieves much better results than the three typical technique analysis and BC.  For the BC,  however, the performances seem quite unstable:  the profit is positive in the IF market while it becomes negative in the  IC market. 
\vspace{0.5em} In addition, two following experiments, including the parameter $k$ in the stop-loss strategy and the trading frequency in the proposed  method,  are also considered. 
 \begin{figure}
    \centering
    \includegraphics[scale=0.58]{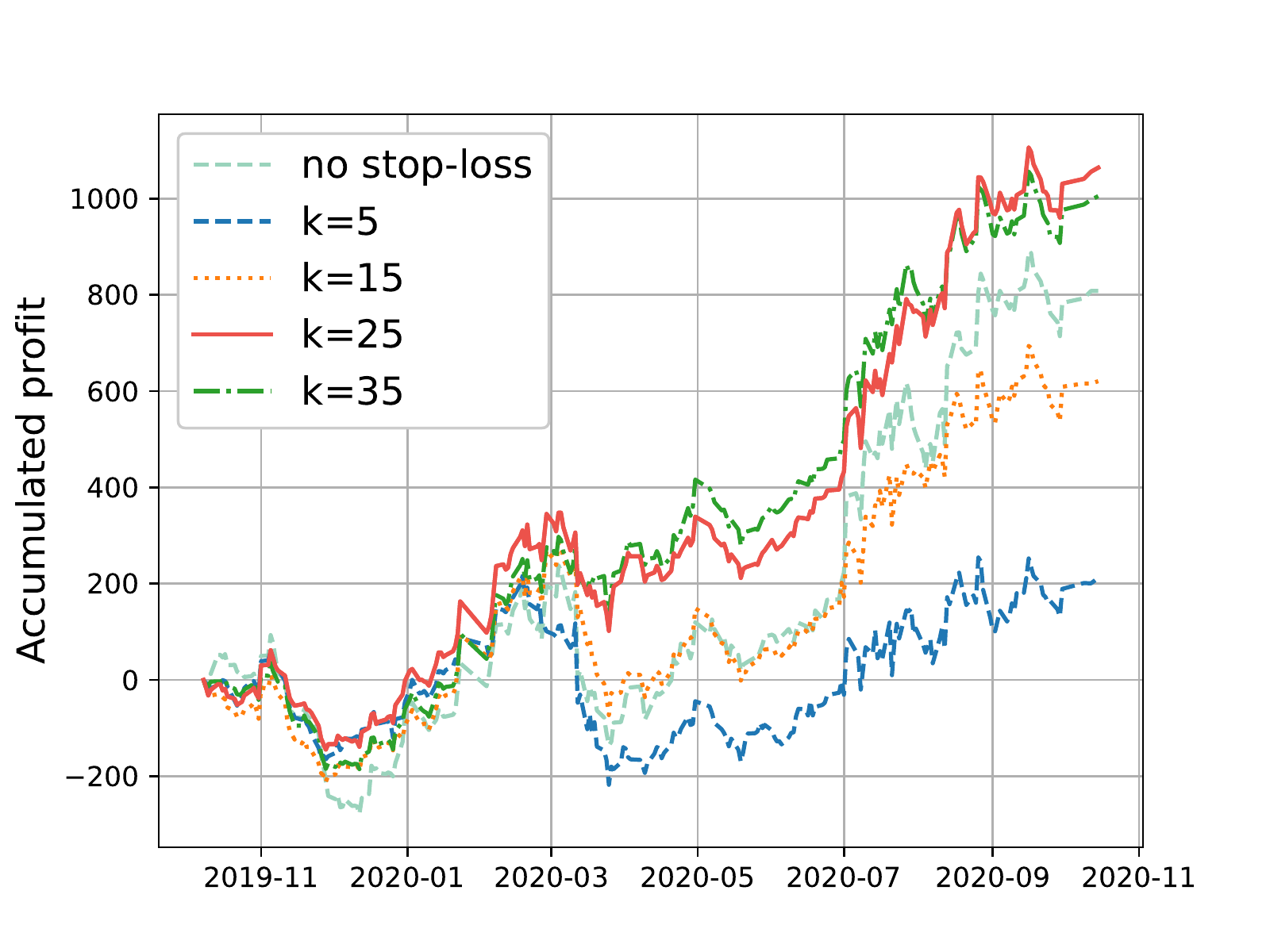}
    \caption{The profit curves with the stop-loss strategy using different parameter $k$ and without using stop-loss strategy. }
    \label{fig:IF_k}
 \end{figure}
\end{itemize}

\vspace{0.5em} \paragraph{About Parameter $k$ in the stop-loss strategy: }  As described in section \ref{subsec:set}, there is a parameter $k$ in the stop-loss strategy.  In this section, we would compare the  profit curves for the stop-loss strategy with  different parameter $k$ ranging from 5 to 35 with a step 10, and without the stop-loss strategy.  The results are shown in Fig. \ref{fig:IF_k}.  From Fig. \ref{fig:IF_k}, we observe that the stop-loss strategy with the parameter $k=25, 35$ can  achieve better profits than others in most cases, and the strategy with $k=25$ finally gains the highest profit in the end.  That is why we set $k=25$ in our experiments.  We also observe that the stop-loss strategy with smaller parameters, e.g., $k=5, 15$,  achieves even poorer results than that without stop-loss strategy in most cases.  Thus, we conclude that the stop-loss strategy is necessary, and its parameter $k$ should be carefully selected.

\vspace{0.5em} \paragraph{About the Trading Frequency:}  In  previous experiments,  we set the trading frequency is 5-min in the proposed method. In this section, we compare the performances for other two  trading frequencies, including 3-min and 30-min.  The  comparative results are  shown in Table \ref{tab:frequency}. From Table \ref{tab:frequency}, we observe that in both IF and IC markets, the proposed method with related higher frequencies (e.g., 3-min and 5-min) works better than that with lower frequency, i.e., 30-min, in our experiments. Taking IC for instance,  the proposed method with 5-min trading frequency gains 1934.3 profits,  while it drops to 405.2 when the  frequency becomes 30-min.  Thus, we conclude that the trading frequency is one of important factors that would affect the proposed method. 
\begin{table}
	\caption{The overall performances of the proposed method with different trading frequencies in IF and IC markets. Those values with an asterisk (*) denote the best results in the corresponding cases.}
	\begin{center}
		\begin{tabular}{c | c | c c c}
            \hline
            \multirow{2}{*}{\textbf{Market}} & \multirow{2}{*}{\textbf{Frequency}} & \multicolumn{3}{c}{\textbf{Performance}}\\
            \textbf{}&\textbf{}&\textbf{Profits}&\textbf{Sharp}&\textbf{Sortino}\\
            \hline
            \multirow{2}{*}{{IF}} &{3min}&892.1&1.76&2.08\\
            \textbf{} & {5min}& \textbf{1064.0 *}&  \textbf{2.09 *}& \textbf{2.47 *}\\
            \textbf{} & {30-min}&763.8&0.37&0.55\\
            \hline
            \multirow{2}{*}{{IC}} &{3min}&1243.1&1.19&1.81\\
            \textbf{} & {5min}& \textbf{1934.3 *}& \textbf{1.86 *}& \textbf{2.86 *}\\
            \textbf{} & {30-min}&405.2&0.21&0.29\\
            \hline
		\end{tabular}
		\label{tab:frequency}
	\end{center}
\end{table}

\section{Conclusion}
\label{sec:Conclusion} 
In this paper, we propose a novel quantitative trading method based reinforcement learning with expert trajectory.  The main contributions of this paper as  follows: 

\begin{itemize}
\item According to the price trend at the next time step,  we design a simple yet very effective expert trajectory.  Agent can effectively learn the optimal policy by utilizing such prior experiences to balance the exploration and exploitation in training process.   

\item  Unlike behavior cloning that  tries to learn the expert's policy using supervised learning,  we first introduce the TD-error generated from both expert-environment interaction and agent-environment interaction to optimize the Q-network for financial applications.   
    
\item Compared with  three typical technique analysis and two RL based methods,   experimental results evaluated two future markets  show that the proposed  method is very promising in quantitative trading. 
\end{itemize}

This is our first attempt to apply the RL with expert trajectory in quantitative trading.   There are many issues worth further studying.  For instance,  we empirically believe that the expert-environment interaction and agent-environment interaction are  equally important for TD-error in the proposed method.  Different weights of the two interactions should be further considered. In addition,  we use a stop-loss strategy to limit the investor's loss on a position.   Some risk measures, such as Sharp ratio and Sortino ratio, should be considered in  reward function and/or loss function in our future works.   
 
\bibliographystyle{splncs04}
\bibliography{arxiv}
\end{document}